\documentclass[10pt,twocolumn,letterpaper]{article}

\usepackage[pagenumbers]{iccv} 

\usepackage{algorithm}
\usepackage{algorithmicx}
\usepackage{makecell}
\usepackage{amssymb}
\usepackage{booktabs}
\usepackage{colortbl}
\usepackage{amsmath}
\usepackage{multirow}
\usepackage{graphicx}
\usepackage{adjustbox}
\usepackage{tikz} 

\usepackage{algpseudocode} 
\usepackage{bm}       

\usepackage{colortbl}
\usepackage{booktabs} 
\usepackage{amsmath}

\definecolor{dbcolor}{rgb}{0,0,1}

\definecolor{mhcolor}{rgb}{1,0,0}

\definecolor{iccvblue}{rgb}{0.21,0.49,0.74}
\usepackage[pagebackref,breaklinks,colorlinks,allcolors=iccvblue]{hyperref}

\def\paperID{13906} 
\def\confName{ICCV}
\def\confYear{2025}

\title{Collaborative Learning for Enhanced Unsupervised Domain Adaptation}

\author{
\textbf{Minhee Cho}\textsuperscript{1},\ 
\textbf{Hyesong Choi}\textsuperscript{1},\ 
\textbf{Hayeon Jo}\textsuperscript{1},\ 
\textbf{Dongbo Min}\textsuperscript{1,*}\\[0.5em]
\textsuperscript{1}Ewha W. University
}

\begin{document}
\maketitle
\begin{abstract}
Unsupervised Domain Adaptation (UDA) endeavors to bridge the gap between a model trained on a labeled source domain and its deployment in an unlabeled target domain. However, current high-performance models demand significant resources, making deployment costs prohibitive and highlighting the need for compact, yet effective models. For UDA of lightweight models, Knowledge Distillation (KD) leveraging a Teacher-Student framework could be a common approach, but we found that domain shift in UDA leads to a significant increase in \textbf{non-salient parameters} in the teacher model, degrading model's generalization ability and transferring misleading information to the student model. Interestingly, we observed that this phenomenon occurs considerably less in the student model. Driven by this insight, we introduce Collaborative Learning for UDA \textbf{(CLDA)}, a method that updates the teacher's non-salient parameters using the student model and at the same time utilizes the updated teacher model to improve UDA performance of the student model. Experiments show consistent performance improvements for both student and teacher models. For example, in semantic segmentation, CLDA achieves an improvement of +0.7\% mIoU for the teacher model and +1.4\% mIoU for the student model compared to the baseline model in the GTA-to-Cityscapes datasets. In the Synthia-to-Cityscapes dataset, it achieves an improvement of +0.8\% mIoU and +2.0\% mIoU for the teacher and student models, respectively.
\end{abstract}    
\section{Introduction}
\label{sec:intro}

\begin{figure}[t]
\centering
\includegraphics[width=1.05 \columnwidth, height=0.7\columnwidth]{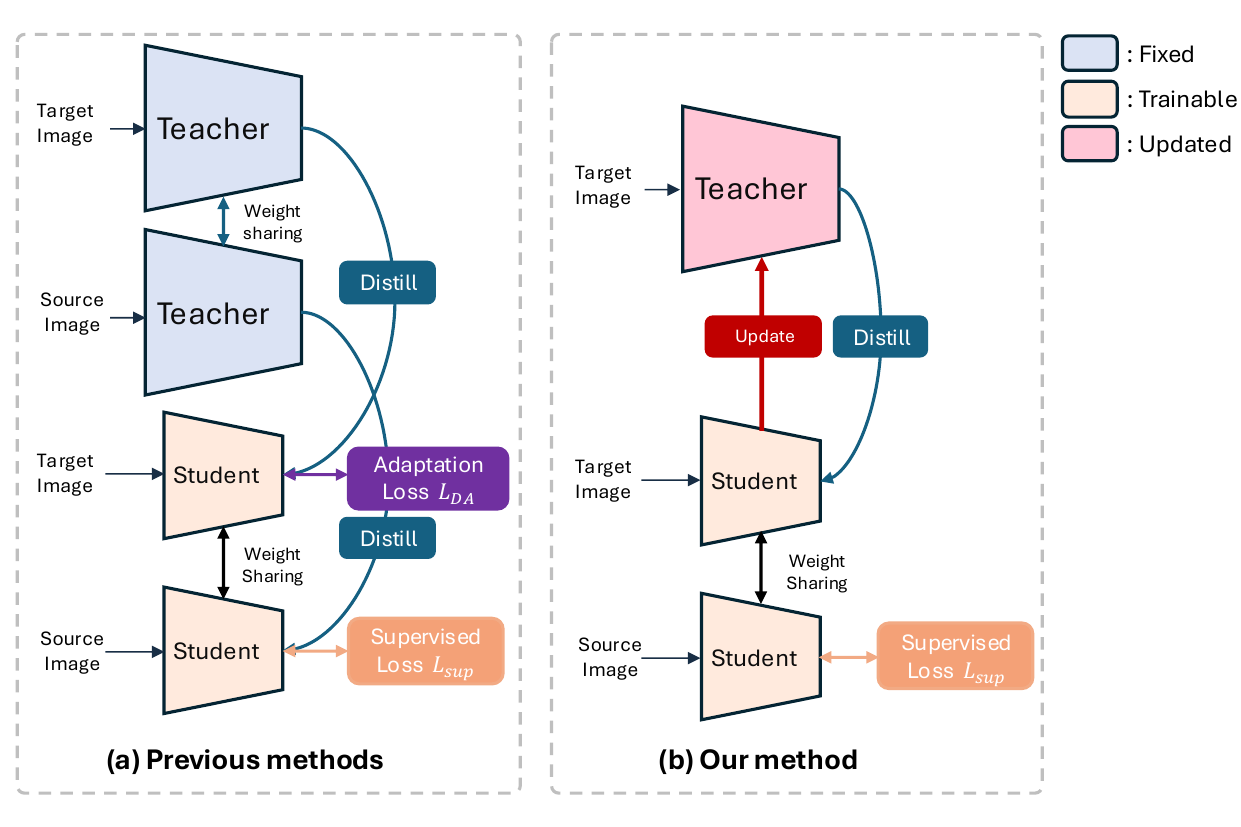} 
\vspace{-25pt}
\caption{ {\bf Conceptual Comparison of UDA Approaches for a Lightweight Model.} While existing KD method for UDA uses a fixed teacher~\cite{kothandaraman2021domain} in (a), our approach updates the teacher through the collaborative learning with the student while allowing the student to fully exploit the enhanced knowledge of the teacher.}
\label{intro:fig1}
\vspace{-0.3cm}
\end{figure}

While the availability of extensive labeled datasets has driven remarkable advancements in various computer vision tasks, there exists a much larger quantity of unlabeled data in real-world scenarios. To accommodate a variety of real-world applications, it is crucial for deep neural networks (DNNs) to generalize to these unlabeled data. However, generalizing the models trained in labeled data (source domain) to unlabeled data (target domain) is non-trivial and pose several challenges. Unsupervised domain adaptation (UDA) has been studied to resolve this issue, which transfers knowledge learned from the source domain to the target domain through adversarial training~\cite{rangwani2022closer,tsai2018learning,toldo2020unsupervised} or self-training~\cite{hoyer2022daformer,hoyer2022hrda,hoyer2023mic}.

Although recent UDA methods~\cite{rangwani2022closer,hoyer2022daformer,hoyer2022hrda} have made remarkable progress, the best-performing approaches are mainly based on resource-intensive networks. Deploying such models in resource-constrained environments is often infeasible. Simply applying conventional UDA methods to lightweight models often fails to maintain competitive performance, underscoring the need for effective UDA strategies tailored to compact models. One possible solution is to apply Knowledge Distillation (KD) that leverages a high-capacity teacher model adapted for the target domain to train a compact student model.  The key in KD is how to effectively transfer knowledge to the student model using the pre-trained teacher model. Building on this key point, recent UDA work~\cite{kothandaraman2021domain} has demonstrated that transferring knowledge from a teacher model improves the generalization ability of a compact student model to the target domain, as shown in Fig.~\ref{intro:fig1} (a).

{However, our preliminary study, which applies the vanilla KD~\cite{hinton2015distilling} to a compact student model using domain-adapted teacher models~\cite{hoyer2022daformer,hoyer2022hrda}, revealed that the approach of using the fixed large teacher model leads to inherent limitations in terms of UDA performance. For a more systematic analysis, we introduce a new metric, termed Layer Saliency Rate (LSR), which measures how salient each layer is to the target domain. This metric aims to quantify the performance degradation induced by the exclusion of a specific layer of the model in the target domain with domain shift. We define the layer with a low LSR value as non-salient, and \emph{vice versa}. We imply that the non-salient layers of the teacher model trained in the source domain do not effectively transfer knowledge to the target domain, thereby exposing limitations in improving generalization performance through the KD.
}

Fig.~\ref{analysis:fig2} analyzes the LSR in UDA based on naive KD~\cite{hinton2015distilling} by training three models: a teacher model (T), a distilled student model (S), and an independently trained student model (IS). The teacher model~(T) and the student model~(IS) trained with the labeled source data are individually adapted to the target domain using self-training based UDA approaches such as DAFormer~\cite{hoyer2022daformer} and HRDA~\cite{hoyer2022hrda}. Then, the knowledge from the teacher model (T) is distilled into the compact student model (S). Notably, the teacher model in this context is not the Mean Teacher model~\cite{tarvainen2017mean}, which has the same size as the student model. Instead, it is a domain-adapted model that is larger than the student model, as in the KD setups. The result indicates that non-salient layers are prevalent in both the teacher model and the independently trained student model.
Notably, this is more pronounced in the teacher model, where over half of the layers are classified as non-salient. This suggests that the pretrained, frozen teacher model (T) is prone to transferring inaccurate information to the student model (S) during KD, and the non-salient layers should be corrected for better knowledge transfer to the student model. We define this case as the problem of \textbf{\underline{D}omain \underline{S}hift induced \underline{N}on-salient parameters (DSN)}.

{Next, we turn our focus to the distilled student model (S), which surprisingly shows the opposite tendency of both the teacher model (T) and the independent student model (IS) in terms of the DSN problem, as shown in Fig.~\ref{analysis:fig2}. This might be because a relatively small number of layers incorporate diverse representations during KD, thus enhancing robustness to domain shifts~\cite{sagawa2020investigation,he2022not}. This implies that the student model has the potential to alleviate the observed DSN problem by correcting the teacher's non-salient layers with the student's layers. Further details are discussed in Section~\ref{subsec:relation}. To the best of our knowledge, our method is the first to explore this problem in the UDA task.} 

Based on this observation, we propose a new UDA approach, {\bf C}ollaborative {\bf L}earning for U{\bf DA}, termed {\bf CLDA}, in which the teacher and student models complement each other to enhance the performance of both models simultaneously. To maximize synergy between the two models, we first identify the non-salient layers in the teacher model where the DSN problem arises, and then establish the layer-wise relations so that we can leverage the corresponding layers of the student model to refine the non-salient layers of the teacher model. As conceptually illustrated in Fig.~\ref{intro:fig1} (b), the non-salient layers of the teacher model are updated by the student model (S $\rightarrow$ T), thereby mitigating the DSN problem and improving the generalization capability in the teacher model. In parallel, the representations of the enhanced teacher model are transferred to the student model (T $\rightarrow$ S), enabling the student to fully exploit this refined information for better adaptation to the target domain. 

In summary, our contributions include the followings; We (1) provide a systematic analysis of the DSN phenomenon in the teacher model, (2) demonstrate that a student model can effectively mitigate the DSN problem in the teacher model, (3) propose a new approach to address the non-salient layers of the teacher model by establishing the layer-wise relations between teacher and student models, and (4) introduce the collaborative learning framework that simultaneously trains both models. 




\section{Related Work}
\label{sec:formatting}

\paragraph{Unsupervised Domain Adaptation.}
Numerous strategies have been proposed to effectively adapt a network to the target domain. These approaches can be categorized into adversarial learning~\cite{gong2019dlow,chen2021scale,long2018conditional} and self-training~\cite{liu2021cycle,zhang2021prototypical,zou2018unsupervised}. Inspired by the success of Generative Adversarial Networks (GANs)~\cite{goodfellow2020generative}, adversarial learning methods aim to learn invariant representations that reduce the distance between source and target distributions at the image~\cite{hoffman2018cycada,kim2020learning}, feature~\cite{tsai2018learning,wang2022cluster,wang2020classes}, and output levels~\cite{vu2019advent,tsai2019domain}. Recently, self-training has emerged as a promising alternative for domain adaptation. Self-training leverages pseudo labels~\cite{lee2013pseudo} for unlabeled target data~\cite{wang2021domain}. To mitigate the noise in pseudo labels caused by domain shift, approaches such as confidence thresholding~\cite{zhang2019curriculum,mei2020instance}, prototypes~\cite{zhang2019category,pan2019transferrable}, and data augmentation~\cite{tranheden2021dacs,french2017self} have been employed. 
To mitigate the noise in pseudo labels caused by domain shift, approaches such as confidence thresholding~\cite{zhang2019curriculum,mei2020instance}, prototypes~\cite{zhang2019category,pan2019transferrable}, and data augmentation~\cite{tranheden2021dacs,french2017self} have been employed. 

\vspace{-0.5cm}
\paragraph{Knowledge Distillation.}
Knowledge distillation (KD) aims to transfer the knowledge acquired by a complex teacher model to a smaller student model. Since~\cite{hinton2015distilling} introduced the concept of knowledge distillation, where the teacher model’s dark knowledge is provided through temperature-scaled softmax outputs, various studies have been inspired to utilize teacher information~\cite{dong2024toward,cho2019efficacy,chen2017learning,luo2016face,heo2019knowledge,park2019relational}. Some methods have been proposed to enhance interaction between the teacher and the student~\cite{park2021learning,dong2024toward}. For instance,~\cite{park2021learning} proposed a novel student-friendly learning technique within the teacher network to facilitate knowledge distillation. However, these methods assume the teacher and student operate within the same domain. When this assumption fails, the teacher's performance degrades, transferring incorrect information to the student. Our research proposes a KD method that remains robust despite domain shifts.


\section{Method}

\begin{figure}[t]
\centering
\includegraphics[width=1 \columnwidth]{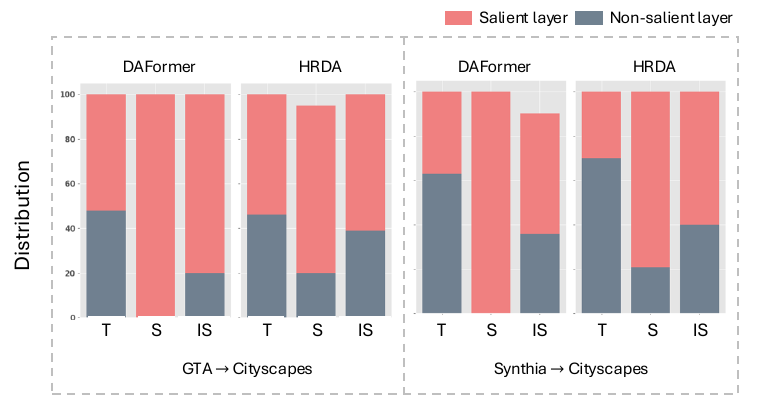} 
\caption{ {\bf The distribution of LSR at the layer level.} {We visualize the distribution of salient and non-salient layers in the fixed teacher model (T), distilled student model (S), and independently trained student model (IS) by measuring the LSR across various UDA methods. Here, the teacher model is a domain-adapted model larger than the student, not a Mean Teacher of the same size.} We evaluated DAFormer~\cite{hoyer2022daformer} and HRDA~\cite{hoyer2022hrda} in domain adaptation scenarios where the source domains are GTA and Synthia, and the target domain is Cityscapes. While more than 50\% of the teacher model's layers suffer from the DSN issue, this problem is significantly less prevalent in the distilled student model.}
\label{analysis:fig2}
\vspace{-0.2cm}
\end{figure}

\subsection{Background and Overview}
The UDA aims to alleviate performance degradation caused by a domain gap between source domain datasets $G=\{ ( x^i_g, y^i_g)\}^{N_G}_{i=1}$ and target domain datasets $Q=\{  ( x^i_q )\}^{N_q}_{i=1}$, where $N_g$ and $N_q$ indicate the number of training images in the source and target domains. A model $f$ comprises a feature extractor $h_{\phi }$ parameterized by $\phi$ and a head $\Phi _{\theta }$ parameterized by $\theta$, \emph{i.e.} $f_{\theta ,\phi } = \Phi _{\theta } \left ( h_{\phi } \left ( x \right ) \right )$. A training process employs a total loss 
consisting of a supervised loss $\mathcal{L}_{sup}$ using the source data $G$ and an domain adaptation loss $\mathcal{L}_{da}$ using the target data $Q$. The type of the supervised loss $\mathcal{L}_{sup}$ varies depending on the tasks~\cite{long2017deep,kan2015bi,hoyer2022daformer}. Also, adaptation loss $L_{da}$ is defined according to the UDA strategy such as adversarial learning~\cite{rangwani2022closer} or self-training \cite{hoyer2022daformer,hoyer2022hrda}. The overall training process is as follows:
\begin{equation}
\min_{\theta,\phi} \frac{1}{N_G} \sum_{k=1}^{N_G} \mathcal{L}_{sup}^{k} + \frac{1}{N_Q} \sum_{k=1}^{N_Q} \mathcal{L}_{da}^{k} \,.
\label{eq:base_loss}
\end{equation}


\noindent As recent UDA methods are based on resource-intensive models, there is a growing need for a lightweight UDA model for deployment in real-world scenarios. Some studies proposed to leverage KD, where a fixed teacher model transfers knowledge to a compact student model during UDA~\cite{kothandaraman2021domain}, as shown in Fig.~\ref{intro:fig1} (a). However, this approach still faces challenges in enhancing UDA performance, as relying on a fixed teacher model can lead to the transfer of misaligned knowledge caused by domain shift. Specifically, the fixed teacher model inherently suffers from the DSN problem, restricting its adaptability to the target domain, whereas the student model is relatively less affected by this problem, as shown in Fig.~\ref{analysis:fig2}. Therefore, to enhance UDA performance more effectively, leveraging the student model is crucial to mitigating the DSN problem in the teacher model.

Instead of mining knowledge from the static teacher, we present a collaborative learning framework in which both teacher and student models are jointly enhanced, as shown in Fig.~\ref{intro:fig1} (b). {\bf (S$\rightarrow$T)}: For the teacher model, we first identify non-salient layers in the teacher model and update them based on the layer-wise relations that are established with the student model. {\bf (T$\rightarrow$S)}: For the student model, we then distill and transfer the refined representations of the updated teacher model to the student model. Fig.~\ref{method:arch} illustrates the overview of the proposed CLDA framework. The layer-wise relation $LR(\gamma)$ and collaborative learning are detailed in Sec.~\ref{subsec:relation} and Sec.~\ref{subsec:CL learning}, respectively. Given the outstanding performance of Transformer-based methods in the UDA, the Transformer-based framework is utilized throughout all experiments.


\subsection{DSN analysis using LSR} \label{subsec:relation}
\paragraph{Layer Saliency Rate (LSR).} 
To quantitatively measure the DSN problem in UDA based on KD~\cite{hinton2015distilling}, we define the LSR that evaluates the relative saliency of each layer in adapting to the target domain. The LSR for a layer $i$ with the parameter $\phi_i$ in the model $f$ is defined as follows:
\begin{equation}
LSR(f,\phi_i) = R(f(\phi ))-R(f(\phi - \phi_i)) \,,
\label{eq:LSR}
\end{equation}
\noindent Here, $R(f(\phi - \phi_i))$ represents the accuracy measured for the target when the layer $i$ is removed, while $R(f(\phi))$ is the accuracy of the original model $f$. A higher LSR value means a higher contribution of the layer. We define the layer with an LSR below a threshold $\tau$ as non-salient. 

\vspace{-0.3cm}
\paragraph{Discussion on Teacher.}
We leverage the LSR to analyze the extent to which the domain-adapted teacher model (T) $f_T$ exacerbates the DSN issue in the presence of domain shift. We quantitatively measure the LSR in the teacher model that is adapted to the target domain using \eqref{eq:base_loss}.
As illustrated in Fig.~\ref{analysis:fig2}, a significant proportion of the teacher model comprises non-salient layers. For instance, in the case of DAFormer~\cite{hoyer2022daformer}, we found that more than 50\% of the layers in the teacher model fall into the non-salient layer. The underlying reason is that larger models tend to overfit to the source domain due to over-parameterization, resulting in many of the learned parameters being invalid or misaligned in the target domain~\cite{sagawa2020investigation,he2022not}. The presence of such non-salient parameters inherently constrains the generalization capability of the teacher model to the target domain. 


\begin{figure}[t]
\centering
\includegraphics[width=0.7 \columnwidth, height=0.8\columnwidth]{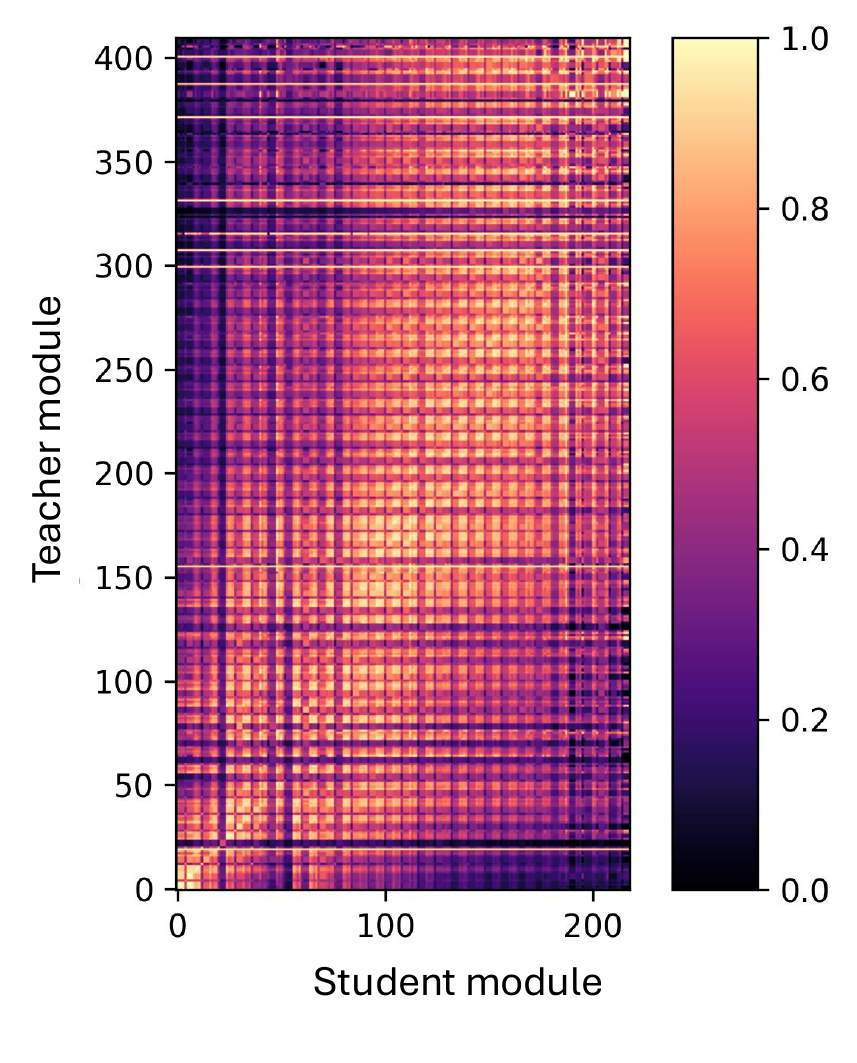} 
\caption{ {\bf CKA Heatmap between Teacher and Distilled Student.} We compute a CKA heatmap between modules within the teacher and distilled student models. The lower half of the student model functionally corresponds to twice the number of modules in the teacher model. Notably, the upper half of the student model aligns with 2.5 times the number of modules in the teacher model.}
\label{analysis:fig3}
\vspace{-0.2cm}
\end{figure}

\vspace{-0.3cm}
\paragraph{Discussion on Student.} 
We extend our analysis to the distilled student model (S) $f_S$, which is trained on the source domain using $L_{sup}$ and on the target domain via vanilla KD~\cite{hinton2015distilling} with the domain-adapted teacher (T). When measuring the LSR using \eqref{eq:LSR}, we observed an opposite phenomenon that the student model (S) exhibits a substantially larger number of salient layers, compared to the teacher model (T) and independently trained student model (IS) in Fig.~\ref{analysis:fig2}. We conjecture that this phenomenon arises from the `condensation' of the roles, which are originally dispersed across multiple layers in the teacher model, into a relatively smaller number of layers of the compact student model, thereby intensifying their generality within the student model’s layers.

For a more principled analysis, we employ a Centered Kernel Alignment (CKA)~\cite{kornblith2019similarity} to measure similarity between deep neural networks at the module level within layers (\eg, attention modules) rather than at the layer level. CKA($X,Y$) $\in $ [0,1] indicates the similarity of two feature vectors $X$ and $Y$. The CKA similarity was measured for all pairs of modules in the layer \( X \in \mathbb{R}^{n \times p_1} \) of the teacher model and \( Y \in \mathbb{R}^{n \times p_2} \) of the distilled student model. The modules with \( p_1 \) and \( p_2 \) dimensions are evaluated using \( n \) examples in the target domain. According to the heatmap results in Fig.~\ref{analysis:fig3}, the lower 0–90 modules of the student model cover the lower 0–180 modules of the teacher model, while the 90–200 modules of the student model cover the remaining parts of the teacher model. This indicates that the student model functionally covers more than 2.5 times the range of the teacher model. This is because in a distilled student model with a relatively small number of layers, layers perform multiple roles, resulting in learning more generalized features and increased robustness in domain-shift situations~\cite{sagawa2020investigation,voita2019analyzing}. This implies that while the non-salient layers in the teacher model suffer from diminished generalization performance on the target domain due to parameters over-optimized for the source domain, the student model can mitigate this limitation by learning more general representations. This finding indicates that the student model can effectively solve the DSN problem of the teacher model in the target domain, beyond simply passively receiving knowledge from the teacher model.
\subsection{CLDA} \label{subsec:CL learning}
Based on our analysis in the previous section, we introduce a collaborative learning framework that takes advantage of the student model to alleviate the DSN problem in the teacher model, and simultaneously enhances the generalization performance of the student model in the target domain through knowledge transfer from the refined teacher model. To put it simply, given the teacher model trained on the source domain and adapted to the target domain, we begin with early KD training stages to stabilize the student model. Subsequently, a layer-wise relation mapping is applied, allowing the student to compensate for the non-salient layers of the teacher model. Based on this mapping, teacher update~(S $\rightarrow$ T) and KD~(T $\rightarrow$ S) are performed alternately to improve the performance of both models simultaneously.

\vspace{-0.3cm}

\paragraph{Layer-wise Relation Mapping.}
We first explore the way of establishing layer-wise relations between the teacher and student models by predicting the most similar layers from the two models, allowing for refining the \emph{non-salient} layers of the teacher model using the corresponding layers of the student model. In the teacher model, the non-salient layers are identified using \eqref{eq:LSR}, and then 30\% of them are randomly selected.
To be specific, we establish layer-wise relations based on the cosine similarity between the feature maps extracted from the teacher model’s non-salient layers and from the student model’s layers. We extract the feature map $A_{T,\gamma} \in \mathbb{R}^{B \times N \times C } $ from the non-salient layer $\gamma$ of the teacher model $f_T$ using the target image. Here, $B,N$ and $C$ represent the batch size, number of tokens, and number of channels, respectively. Similarly, we extract the feature map $ \left\{A_{S,i}\right\}_{i=1}^{L_{s}} \in \mathbb{R}^{B \times N \times C } $ from the same target image across all layers $L_{s}$ within the student model.


Then, in the training step $\eta$, we select the layer of the student model that has the highest cosine similarity between the channels in the feature maps of the teacher and student models as follows:
\begin{equation}
LR(\gamma) = \underset{i}{\text{argmax}} \sum_{\eta=T_0}^{T_{LR}}\ \vartheta^{\eta}_{i}\,,
\label{eq:layer_index}
\end{equation}

\noindent where the cosine similarity $\vartheta$ is defined as
\begin{equation}
   \vartheta^{\eta}_{i} = \sum_{j=1}^{C}\frac{A_{T,\gamma}^{n}[:, : , j] \cdot A_{S,i}^{n}[:, : , j]}{|A_{T,\gamma}^{n}[:, : , j]| \cdot |A_{S,i}^{n}[:, : , j]|}\,.
\label{eq:layer_mapping}
\end{equation}

\noindent Since the representations across layers in the student model are initially unstable during early training stages, the layer-wise relation $LR$ with the teacher model is established with the feature maps between early KD training stages $T_0$ and $T_{LR}$.

\begin{figure}[t]
\centering
\includegraphics[width=1 \columnwidth,height=0.8 \columnwidth]{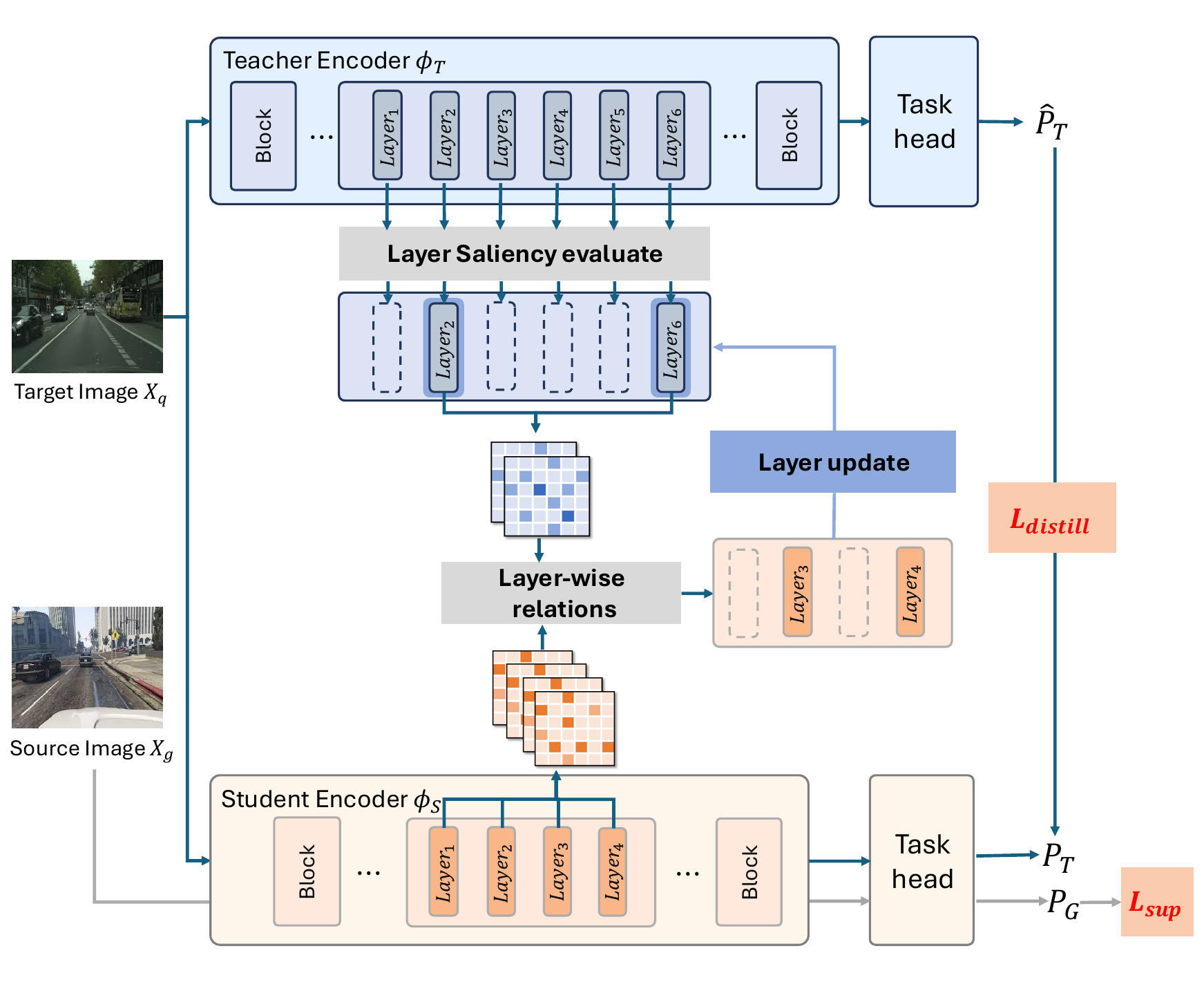}

\caption{ {\bf An illustration of the proposed CLDA framework.} For the teacher model, we first identify DSN layers and establish layer-wise relations with the student model. Based on the layer-wise relations, we update non-salient layers to mitigate their DSN problem. The student model then incorporates the refined representations from the updated teacher model, leveraging enhanced generalization to improve performance in the target domain. }
\label{method:arch}
\vspace{-0.3cm}
\end{figure}

\vspace{-0.4cm}

\paragraph{Teacher Update (S$\rightarrow$T).}
Recent UDA methods~\cite{hoyer2022daformer,hoyer2022hrda,hoyer2023mic,wang2023cdac} leveraging Transformer architectures maintain a consistent channel dimension across blocks, enabling efficient layer-wise updates. Exploiting this property, we design the teacher update strategy based on the layer-wise relations $LR(\gamma)$. To be specific, we employ an exponential moving average (EMA) strategy, progressively integrating the student model’s parameters into the teacher’s non-salient layers. This facilitates effective collaborative learning without additional backpropagation overhead in the teacher model. At training step $\eta$, the teacher parameter is updated with the corresponding student parameter:
\begin{equation}
\phi_{T,\gamma}^{\eta} \leftarrow \alpha \phi_{T,\gamma}^{\eta } + (1 - \alpha) \phi_{S,LR(\gamma)}^{\eta } \,,
\label{eq:ema}
\end{equation}
\noindent where $\phi_{T,\gamma}^{\eta}$ indicates the teacher parameter at layer $\gamma$, and $\phi_{S,LR(\gamma)}^{\eta }$ is the corresponding student parameter at layer $LR(\gamma)$.

\paragraph{Knowledge Distillation (T$\rightarrow$S).}
The student model learns from the source domain using the supervised loss $L_{sup}$, while for the target domain, it acquires knowledge from the teacher model trained via \eqref{eq:base_loss} as in recent UDA approaches~\cite{hoyer2022daformer,hoyer2022hrda,hoyer2023mic}. The KD process is performed in the output space using $L_{Distill}$;
\begin{equation}
\mathcal{L}_{distill}^{(i)} = - \sum_{j=1}^{H \times W} \sum_{c=1}^{N_{C}} q^{i} \hat{p}_T^{(i,j,c)} \log f_{S}\left(x_q^{i}\right)^{(j,c)}\,.
\label{eq:distill_loss}
\end{equation}


\noindent where $H$ and $W$ are height and width of an target image $x_q$, respectively, while $N_C$ is the number of categories shared between the source and the target domain. Hard pseudo labels $\hat{p}_T$ are generated from most confident class prediction in the teacher model.
 \begin{equation}
\hat{p}_T^{(i,j,c)} = \left[ c = \arg\max_{c'} f_{T}\left(x_q^{(i)}\right)^{(j,c')} \right]\,.
\label{eq:PGT}
\end{equation}

\noindent where $\left [  \cdot \right ]$ denotes the Iverson bracket. Since the hard pseudo labels may contain potential errors, the loss is weighted by the quality estimate $q$ of the pseudo label~\cite{hoyer2022hrda,hoyer2022daformer,zhang2018collaborative}.  


\paragraph{Total Loss.}
In the early KD training stages $T_0 $, we first apply $ L_{\text{distill}} $ to alleviate the initial instability of the student model in the target domain. After $T_{LR}$, teacher update and distillation are alternately performed during training. The total loss of the proposed CLDA is formulated as
\vspace{-0.3cm}

\begin{equation}
L_{CLDA} =  \min_{\theta,\phi} \frac{1}{N_G} \sum_{k=1}^{N_G} \mathcal{L}^k_{sup} + \frac{1}{N_Q} \sum_{k=1}^{N_Q} \mathcal{L}^k_{distill} \,.
\label{eq:our_loss}
\end{equation}
\noindent The overall algorithm is summarized in supplementary Sec.~4
\section{Experiments}
\label{sec:experiments}

\begin{table*}[!t]
\caption{Comparison of the class-wise IoU of existing semantic segmentation UDA methods with our proposed CLDA in the GTA-to-Cityscapes datasets.}
\vspace{-0.2cm}
\centering
\addtolength{\tabcolsep}{-4pt}
\resizebox{\textwidth}{!}{
{\small
\begin{tabular}{c|ccccccccccccccccccc|c}
\toprule
\hline
\multicolumn{21}{c}{GTA $\to$ Cityscapes} \\ \toprule
\hline
\multicolumn{1}{l|}{Method} & Road & S.walk & Build & Wall & Fence & Pole & Tr.Light & Tr.Sign & Veget & Terrain & Sky  & Person & Rider & Car  & Truck & Bus  & Train & M.Bike &  \multicolumn{1}{l|}{Bike} & mIoU(\%) \\ 
\hline
\multicolumn{1}{l|}{ADVENT~\cite{vu2019advent}} & 89.4 & 33.1 & 81.0 & 26.6 & 26.8 & 27.2 & 33.5 & 24.7 & 83.9 & 36.7 & 78.8 & 58.7 & 30.5 & 84.8 & 38.5 & 44.5 & 1.7 & 31.6 & \multicolumn{1}{l|}{32.4 } &  45.5\\ 
\multicolumn{1}{l|}{ProDA~\cite{zhang2021prototypical}} & 87.8 & 56.0 & 79.7 & 46.3 & 44.8 & 45.6 & 53.5 & 53.5 & 88.6 & 45.2 & 82.1 & 70.7 & 39.2 & 88.8 & 45.5 & 59.4 & 1.0 & 48.9 & \multicolumn{1}{l|}{56.4 } &  57.5\\
\multicolumn{1}{l|}{CaCo~\cite{huang2022category}} & 93.8 & 64.1 & 85.7 & 43.7 & 42.2 & 46.1 & 50.1 & 54.0 & 88.7 & 47.0 & 86.5 & 68.1 & 2.9 & 88.0 & 43.4 & 60.1 & 31.5 & 46.1 &  \multicolumn{1}{l|}{60.9} & 58.0 \\
\multicolumn{1}{l|}{ADPL~\cite{cheng2023adpl}} & 93.4 & 60.6 & 87.5 & 45.3 & 32.6 & 37.3 & 43.3 & 55.5 & 87.2 & 44.8 & 88 & 64.5 & 34.2 & 88.3 & 52.6 & 61.8 & 49.8 & 41.8 & \multicolumn{1}{l|}{59.4 } &  59.4\\ 
\hline
\multicolumn{1}{l|}{DAFormer [MiT-B5]~\cite{hoyer2022daformer} } & 95.7 & 70.2 & 89.4 & 53.5 & 48.1 & 49.6 & 55.8 & 59.4 & 89.9 & 47.9 & 92.5 & 72.2 & 44.7 & 92.3 & 74.5 & 78.2 & 65.1 & 55.9 &  \multicolumn{1}{l|}{61.8} & 68.3 \\

\rowcolor[rgb]{0.9, 0.9, 0.9} \multicolumn{1}{l|}{CLDA (DAFormer [T]) } & 96.5 & 73.8 & 89.5 & 53.6 & 49.2 & 50.3 & 54.5 & 63.4 & 89.9 & 45.8 & 92.4 & 71.8 & 44.8 & 92.4 & 77.4 & 80.5 & 67.8 & 54.3 &  \multicolumn{1}{l|}{63.3} & {\bf 69.0} \\

\multicolumn{1}{l|}{DAFormer [MiT-B3]~\cite{hoyer2022daformer}} & 96.8 &  75.4 &  89.2 &  52.8 & 43.6 & 49.5 & 55.7 &  62.2 & 89.9 &  47.8 & 90.8 & 71.4 &  43.3 &  91.5 &  66.1 &  76.3 &  72.0 &  55.5 &  \multicolumn{1}{l|}{ 63.4} & 68.1\\ 

\rowcolor[rgb]{0.9, 0.9, 0.9} \multicolumn{1}{l|}{CLDA (DAFormer [S) } & 96.8 & 75.2 & 89.4 & 50.6 & 48.7 & 50.5 & 55.5 & 63.9 & 89.9 & 45.6 & 92.7 & 72.3 & 45.7 & 92.8 & 78.9 & 80.9 & 69.9 & 54.7 &  \multicolumn{1}{l|}{63.7} & {\bf 69.5} \\
\hline
\multicolumn{1}{l|}{HRDA [MiT-B5]~\cite{hoyer2022hrda} } & 96.4 & 74.4 & 91.0 & 61.6 & 51.5 & 57.1 & 63.9 & 69.3 & 91.3 & 48.4 & 94.2 & 79.0 & 52.9 & 93.9 & 84.1 & 85.7 & 75.9 & 63.9 &  \multicolumn{1}{l|}{67.5} & 73.8 \\
\rowcolor[rgb]{0.9, 0.9, 0.9} \multicolumn{1}{l|}{CLDA (HRDA [T]) } & 96.7 & 75.8 & 91.3 & 59.9 & 54.5 & 58.6 & 65.1 & 70.3 & 91.7 & 51.4 & 94.5 & 79.3 & 53.3 & 94.1 & 84.6 & 87.0 & 76.9 & 65.1 &  \multicolumn{1}{l|}{68.2} & { \bf 74.6 } \\
\multicolumn{1}{l|}{HRDA [MiT-B3]~\cite{hoyer2022hrda}  } &  95.9 &  72.7 &  90.9 &  55.3 & 48.9 & 59.0 & 64.9 &  72.2 & 91.3 &  50.4 & 93.3 & 77.7 & 50.1 &  93.5 & 83.4 &  84.0 &  75.4 &  62.8 &  \multicolumn{1}{l|}{66.2} & 73.0\\ 
\rowcolor[rgb]{0.9, 0.9, 0.9} \multicolumn{1}{l|}{CLDA (HRDA [S]) } & 96.8 & 76.5 & 91.2 & 60.3 & 55.8 & 57.7 & 64.8 & 70.0 & 91.5 & 50.7 & 94.3 & 79.1 & 52.6 & 94.2 & 85.3 & 85.6 & 73.9 & 64.9 &  \multicolumn{1}{l|}{67.4} & {\bf 74.4} \\

\hline
\end{tabular}}}
\label{tab:gta} 
\vspace{-0.3cm}
\end{table*}


\begin{table*}[!t]
\caption{Comparison of the class-wise IoU of existing semantic segmentation UDA methods with our proposed CLDA in the Synthia-to-Cityscapes datasets.}
\vspace{-0.2cm}
\centering
\addtolength{\tabcolsep}{-4pt}
\resizebox{\textwidth}{!}{
{\small
\begin{tabular}{c|ccccccccccccccccccc|c}
\toprule
\hline
\multicolumn{21}{c}{Synthia $\to$ Cityscapes} \\ \toprule
\hline
\multicolumn{1}{l|}{Method} & Road & S.walk & Build & Wall & Fence & Pole & Tr.Light & Tr.Sign & Veget & Terrain & Sky  & Person & Rider & Car  & Truck & Bus  & Train & M.Bike &  \multicolumn{1}{l|}{Bike} & mIoU(\%) \\ 
\hline
\multicolumn{1}{l|}{ADVENT~\cite{vu2019advent}} & 85.6 & 42.2 & 79.7 & 8.7 & 0.4 & 25.9 & 5.4 & 8.1 & 80.4 & - & 84.1 & 57.9 & 23.8 & 73.3 & - & 36.4 & - & 14.2 &  \multicolumn{1}{l|}{33.0} & 41.2 \\
\multicolumn{1}{l|}{ProDA~\cite{zhang2021prototypical}} & 87.8 & 45.7 & 84.6 & 37.1 & 0.6 & 44.0 & 54.6 & 37.0 & 88.1 & - & 84.4 & 74.2 & 24.3 & 88.2 & - & 51.1 & - & 40.5 & \multicolumn{1}{l|}{45.6 } &  55.5\\ 
\multicolumn{1}{l|}{CaCo~\cite{huang2022category} } & 87.4 & 48.9 & 79.6 & 8.8 & 0.2 & 30.1 & 17.4 & 28.3 & 79.9 & - & 81.2 & 56.3 & 24.2 & 78.6 & - & 39.2 & - & 28.1 &  \multicolumn{1}{l|}{48.3} & 46.0 \\
\multicolumn{1}{l|}{ADPL~\cite{cheng2023adpl}} & 86.1 & 38.6 & 85.9 & 29.7 & 1.3 & 36.6 & 41.3 & 47.2 & 85 & - & 90.4 & 67.5 & 44.3 & 87.4 & - & 57.1 & - & 43.9 & \multicolumn{1}{l|}{51.4 } &  55.9\\ 
\hline
\multicolumn{1}{l|}{DAFormer [MiT-B5]~\cite{hoyer2022daformer} } & 84.5 & 40.7 & 88.4 & 41.5 & 6.5 & 50.0 & 55.0 & 54.6 & 86.0 & - & 89.8 & 73.2 & 48.2 & 87.2 & - & 53.2 & - & 53.9 &  \multicolumn{1}{l|}{61.7} & 60.9 \\
\rowcolor[rgb]{0.9, 0.9, 0.9}\multicolumn{1}{l|}{CLDA (DAFormer [T]) } & 85.1 & 42.5 & 87.9 & 42.2 & 7.1 & 50.5 & 55.5 & 55.4 & 86.3 & -  & 89.4 & 71.8 & 48.6 & 87.8 &  -  & 62.2 &  - & 53.5 &  \multicolumn{1}{l|}{61.3} & {\bf 61.7} \\
\multicolumn{1}{l|}{DAFormer [MiT-B3]~\cite{hoyer2022daformer} } &  87.0 &  45.9 &  88.1 &  40.3 & 3.7 & 48.5 & 53.4 &  52.4 & 86.7 &  -  & 88.8 & 73.3 &  42.2 &  85.8 &  - &  58.3 &  - &  46.1 &  \multicolumn{1}{l|}{ 49.9} & 59.4\\ 
\rowcolor[rgb]{0.9, 0.9, 0.9}\multicolumn{1}{l|}{CLDA (DAFormer [S]) } & 85.4 & 42.8 & 87.9 & 41.6 & 7.2 & 50.1 & 56.0 & 55.7 & 86.4 & - & 89.8 & 72.2 & 48.8 & 87.4 & - & 56.3 & - & 52.9 &  \multicolumn{1}{l|}{61.4} &{ \bf 61.4} \\
\hline
\multicolumn{1}{l|}{HRDA [MiT-B5]~\cite{hoyer2022hrda} } & 85.2 & 47.7 & 88.8 & 49.5 & 4.8 & 57.2 & 65.7 & 60.9 & 85.3 & -  & 92.9 & 79.4 & 52.8 & 89.0 & - & 64.7 & - & 63.9 &  \multicolumn{1}{l|}{64.9} & 65.8 \\
\rowcolor[rgb]{0.9, 0.9, 0.9}\multicolumn{1}{l|}{CLDA (HRDA[T]) } & 85.1 & 52.5 & 89.5 & 47.6 & 7.0 & 58.9 & 66.2 & 62.9 & 82.2 & - & 94.0 & 80.0 & 52.3 & 87.2 & - & 67.6 & - & 62.0 &  \multicolumn{1}{l|}{53.1} & { \bf 66.1 } \\
\multicolumn{1}{l|}{HRDA [MiT-B3]~\cite{hoyer2022hrda}  } &  87.4 &  51.9 &  89.3 &  48.7 & 2.4 & 58.7 & 65.6 &  57.4 & 85.0 &  - & 93.8 & 77.9 &  51.6 &  87.3 &  - &  66.0 &  - &  61.1 &  \multicolumn{1}{l|}{65.9} &  65.7\\ 
\rowcolor[rgb]{0.9, 0.9, 0.9} \multicolumn{1}{l|}{CLDA (HRDA[S])} & 92.2 & 70.1 & 89.6 & 49.6 & 2.6 & 59.1 & 65.3 & 60.2 & 83.8 & - & 93.7 & 79.1 & 52.0 & 86.6 & - & 67.7 & - & 59.1 &  \multicolumn{1}{l|}{65.5} & { \bf 67.2 } \\

\hline
\end{tabular}}}
\label{tab:synthia} 
\vspace{-0.3cm}
\end{table*}


\subsection{Implementation Details}

\paragraph{Semantic Segmentation:}
Our approach was applied to various UDA frameworks such as DAFormer~\cite{hoyer2022daformer} and HRDA~\cite{hoyer2022hrda}. Specifically, we followed the configuration of the DAFormer, where the teacher model [T] employs a MiT-B5 encoder~\cite{xie2021segformer} of 81.4M parameters pre-trained on ImageNet~\cite{deng2009imagenet}, while the student model [S] utilizes a MiT-B3 encoder~\cite{xie2021segformer} of 44M parameters.  AdamW~\cite{loshchilov2017decoupled} was used as an optimizer, with a learning rate of $6 \times 10^{-5}$ for the encoder, $6 \times 10^{-4}$ for the decoder, a weight decay of 0.01, and a batch size of 2. During the training phase, we applied DACS~\cite{tranheden2021dacs} data augmentation as in \cite{hoyer2022daformer}. The EMA coefficient $ \alpha $ in \eqref{eq:ema} is set to 0.9999 and threshold $\lambda$ for \eqref{eq:LSR} is set to 0.1. In the layer-wise relation mapping process, $T_0$ and $T_{LR}$ were set to 2K and 2.5K iterations, respectively. As source data, GTA~\cite{richter2016playing} is a synthetic urban scene dataset, containing 24,966 images with pixel-level annotations, each with a resolution of 1914 $\times$ 1052. Synthia~\cite{ros2016synthia} is another synthetic urban scene dataset, which includes 9,400 images and corresponding annotations at a resolution of 1280 $\times$760. As target data, Cityscapes~\cite{cordts2016cityscapes} is the real-world urban scene dataset that includes 2,975 training images and 500 holdout images for evaluation, with an image resolution of 2048 $\times$ 1024. The training resolution for each dataset follows the UDA methods~\cite{hoyer2022hrda,hoyer2022daformer} used as baselines; Specifically, DAFormer~\cite{hoyer2022daformer} operates at half resolution, while HRDA~\cite{hoyer2022hrda} uses the full resolution.

\paragraph{Image Classification:} We evaluated adaptation performance in the image classification task using the VisDA-2017 dataset~\cite{peng2017visda}, which includes 280,000 synthetic and real images across 12 distinct classes. Our experiments utilized a teacher-student framework where the teacher model [T] is a ViT-L/16 and the student model [S] is a ViT-B/16~\cite{dosovitskiy2020image}. For UDA, we employ the Smooth Domain Adaptation Technique (SDAT)~\cite{rangwani2022closer}, which leverages the Conditional Domain Adaptation Network (CDAN)~\cite{long2018conditional} and the Margin Consistency Criterion (MCC)~\cite{jin2020minimum}. The training was conducted using Stochastic Gradient Descent (SGD), with a learning rate of 0.002 and a batch size of 32.



\begin{table*}[h]
\caption{Image classification accuracy in \% on VisDA-2017~\cite{peng2017visda} for UDA. The last column contains the mean across classes.}
\vspace{-3pt}
\centering
\addtolength{\tabcolsep}{-4pt}
\resizebox{0.75\textwidth}{!}{
{\small
\begin{tabular}{l|ccccccccccccccc|c}
\toprule
\multicolumn{1}{l|}{Method} & Plane & Bcycl & Bus & Car & Horse & Knife & Mcyle & Persn & Plant & Sktb & Train & Truck & Mean \\ 
\midrule

\multicolumn{1}{l|}{TVT~\cite{yang2023tvt} } & 92.9 & 85.6 & 77.5 & 60.5 & 93.6 & 98.2 & 89.3 & 76.4 & 93.6 & 92.0 & 91.7 & 55.7 & 83.9 \\
\multicolumn{1}{l|}{CDTrans~\cite{xu2021cdtrans} } & 97.1 & 90.5 & 82.4 & 77.5 & 96.6 & 96.1 & 93.6 & 88.6 & 97.9 & 86.9 & 90.3 & 62.8 & 88.4 \\
\hline
\multicolumn{1}{l|}{SDAT [ViT-L]~\cite{rangwani2022closer} } & 98.8 & 91.3 & 85.6 & 77.3 & 98.5 & 97.4 & 96.0 & 83.0 & 95.5 & 98.3 & 94.5 & 68.4 & 90.4 \\

\rowcolor[rgb]{0.9, 0.9, 0.9} \multicolumn{1}{l|}{CLDA (SDAT [T])} & 99.0 & 91.1 & 86.9 & 78.8 & 99.0 & 98.6 & 96.8 & 86.0 & 94.8 & 97.5 & 95.4 &70.5 & \textbf{91.2} \\

\multicolumn{1}{l|}{SDAT [ViT-B]~\cite{rangwani2022closer}} & 98.4 & 90.9 & 85.4 & 82.1 & 98.5 & 97.6 & 96.3 & 86.1 & 96.2 & 97.6 & 92.9 & 56.8 & 89.8 \\ 

\rowcolor[rgb]{0.9, 0.9, 0.9} \multicolumn{1}{l|}{CLDA (SDAT [S])} & 98.7 & 92.3 & 85.0 & 73.8 & 98.6 & 98.2 & 96.1 & 84.4 & 96.9 & 98.1 & 93.7 & 65.1 & \textbf{90.1} \\
\bottomrule
\end{tabular}}}
\label{tab:sdat} 
\vspace{-0.3cm}
\end{table*}
\subsection{Comparative Study}
\paragraph{Semantic Segmentation:}
CLDA was integrated with recent UDA methods. Table~\ref{tab:gta} and \ref{tab:synthia} demonstrate that CLDA consistently enhances performance in different datasets, especially confirming improvements in both the teacher and student models. When employing DAFormer on the GTA-to-Cityscapes dataset, the teacher and student models exhibited a performance improvement of 0.7\% and 1.4\%, respectively. In addition, using HRDA, the teacher and student models achieved performance improvements of 0.8\% and 1.4\%, respectively. Similarly, on the Synthia-to-Cityscapes dataset, CLDA improved performance by 0.8\% for the teacher and 2.0\% for the student with DAFormer, and by 0.3\% for the teacher and 1.5\% for the student with HRDA. These findings suggest that CLDA can be flexibly combined with various Transformer-based models. Notably, our approach improves the teacher model without additional backpropagation of the teacher, underscoring the importance of transmitting informative knowledge to the student model and maximizing the synergy between the two models. In the student, our approach enabled the compact student model to achieve performance comparable to that of the teacher model.


\paragraph{Image Classification:}
For the evaluation in image classification, we adopted the recently proposed SDAT~\cite{rangwani2022closer}, which is trained through adversarial learning. As shown in Tab.~\ref{tab:sdat}, our evaluation on the VisDA-2017 dataset showed that CLDA enhanced the UDA performance by +0.8\% and +0.3\% for the teacher and student models, demonstrating that CLDA is effective not only in the self-training approach but also in the adversarial learning-based UDA method. This suggests that CLDA has the generality to be applied effectively to various methods in the UDA task. In particular, CLDA plays an important role in simultaneously improving the performance of teacher and student models through collaborative learning.

\begin{table}[t]
    \caption{Ablation Study on Components of CLDA.}
    \centering
    \begin{tabular}{ccccc|c}
        \toprule
        \multicolumn{2}{c|}{Base} & \multicolumn{1}{c|}{T$\rightarrow$S} & \multicolumn{2}{c|}{S$\rightarrow$T} & \\ 
        \cmidrule(r){1-5}
        $L_s$ & $L_{da}$ & $L_{distill}$ & \textit{random} & $R_{layer}$ & \textbf{mIoU(\%)} \\
        \midrule
        \checkmark & \checkmark & & & & 68.1 \\
        \checkmark &  & \checkmark & & & 68.9 \\
        \checkmark &  & \checkmark & \checkmark & & 68.3 \\
        \checkmark & & \checkmark & & \checkmark & 69.5 \\
        \bottomrule
    \end{tabular}
    \label{table:Ablation1}
\vspace{-0.3cm}
\end{table}


\subsection{Ablation study of CLDA}
\begin{table}[t]
\centering
\caption{The mIoU~(\%) of different numbers of layers used for mapping layer-wise relations.}
\label{tab:relation1}
\renewcommand{\arraystretch}{1.2}
\small\addtolength{\tabcolsep}{5pt}
\begin{tabular}{l|c|>{\columncolor{gray!20}}c|c}
\toprule
\textbf{} & \textbf{10\%} & \textbf{30\%} & \textbf{50\%} \\
\midrule
Teacher & 68.8 & 69.0 & 68.5 \\
Student & 69.4 & 69.5 & 69.1 \\
\bottomrule
\end{tabular}
\vspace{-0.3cm}
\end{table}

\paragraph{Analysis on Individual Component:} 
In Table~\ref{table:Ablation1}, we conducted a component ablation study on the GTA$\rightarrow$Cityscape dataset with DAFormer using MiT-b3. The complete CLDA configuration achieves the mIoU of 69.5 (row 4), which is +1.4\% mIoU higher than DAFormer independently trained without KD (row 1). When distilling from the teacher to the student (T$\rightarrow$S) in row 2, the performance improvement of 0.8\% was achieved over the DAFormer. However, when updating teacher layers by randomly selecting student layers without layer-wise relation mapping in row 3, the mismatch between student and teacher layers led to a performance degradation of 0.6\%, when compared to the KD in row 2. On the other hand, when the layer-wise relation mapping was established before updating the teacher layers, an improvement of 0.6\% was observed. The results suggest that during the CLDA training process, it is crucial for the teacher to establish corresponding layers with the student layers. Furthermore, we observed that when the student model inherits knowledge from an updated teacher model constructed based on well-defined layer-wise relations, its generalization capability improves, leading to enhanced performance in the target domain.


\vspace{-0.6cm}
\paragraph{Layer-wise Relations Mapping:}
We conducted an ablation study on building the layer-wise relation on the GTA$\rightarrow$Cityscapes. Table~\ref{tab:relation1} analyzes the impact of the number of layers in which the layer-wise relation mapping is applied. The highest performance (69.5 mIoU) was achieved when it was set to 30\%. However, increasing it to 50\% led to 0.5\% performance drop, with the student model experiencing an additional 0.4\% degradation. This decline can be attributed to the limited number of layers in the student model, which could not effectively correspond with the non-salient layers in the teacher model. 
Table~\ref{tab:relation2} investigates the impact of threshold $\tau$ to determine non-salient and salient layers in \eqref{eq:LSR}. When the threshold exceeded 0.1, the distinction between salient and non-salient layers became ambiguous, making it difficult to accurately identify non-salient layers. This underscores the importance of selecting an appropriate threshold to ensure effective layer-wise relation mapping.

\begin{table}[t]
\centering
\caption{The mIoU~(\%) of different thresholds in layer-wise relation mapping.}
\label{tab:relation2}
\renewcommand{\arraystretch}{1.2}
\small\addtolength{\tabcolsep}{5pt}
\begin{tabular}{l|>{\columncolor{gray!20}}c|c|c|c}
\toprule
\textbf{} & \textbf{0.1} & \textbf{0.3} & \textbf{0.5} & DAFormer \\
\midrule
Teacher & 69.0 & 68.2 & 68.1 & 68.3 \\
Student & 69.5 & 69.2 & 69.0 & 68.1 \\
\bottomrule
\end{tabular}
\vspace{-0.5cm}
\end{table}

\subsection{Discussion}
\begin{figure}[t]
\centering
\includegraphics[width=0.9\columnwidth]{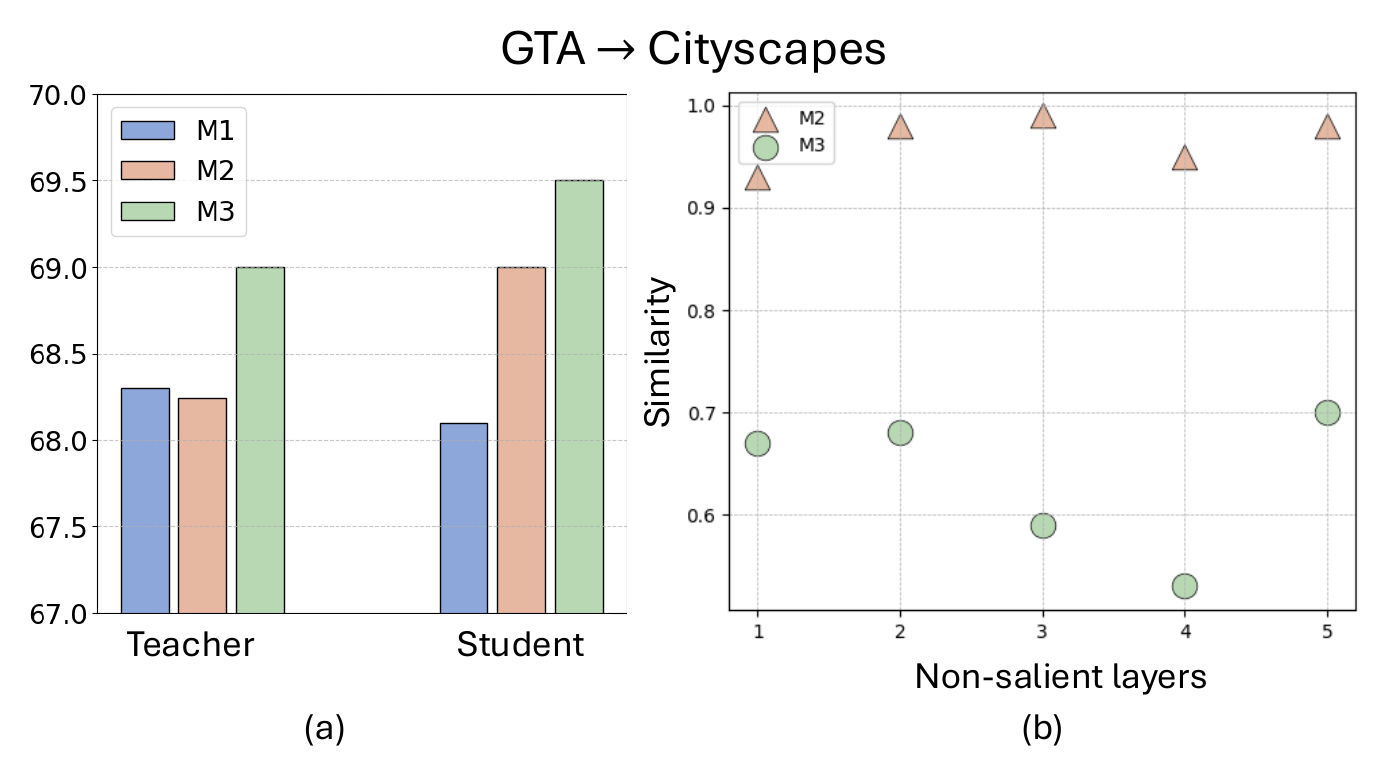} 
\vspace{-0.3cm}
\caption{ {\bf Comparison on Student Model Size in CLDA.} While `M1' indicates baseline results where the teacher (MiT-B5) and student (MiT-B3) models are trained individually, `M2' are `M3' represents the results obtained by applying CLDA. In `M2', the student model is the same size as the teacher (MiT-B5). `M3' indicates the original setup of CLDA, where a teacher model (MiT-B5) and a smaller student model (MiT-B3) are used. (a) Comparison of performance based on student model size. (b) Similarity between non-salient teacher layers and compact student layers, as well as the layer similarity within the student model of the same size.}
\label{experiments:fig5}
\vspace{-0.5cm}
\end{figure}

\paragraph{Impact of student.} 
To verify that the key contribution to resolving the teacher model’s DSN issue stems from the student model, we pose the following question: ``If the student model shares the same architecture as the teacher, similar to the standard Mean Teacher framework~\cite{tarvainen2017mean}, can we still improve teacher performance by updating the teacher model with the help of the student model?" Experimental results for GTA5$\rightarrow$Cityscapes dataset in Fig.~\ref{experiments:fig5} (a) indicate that when the student model (MiT-B5) was used identical to the teacher model~`M2', the UDA performance in the student model was reduced to 69.0\%, which was even lower than using a compact student model (MiT-B3) of 69.5\% in `M3'. Notably, the teacher’s performance remained at 68.2\%, showing no improvement over the baseline in `M1' where the teacher and student models were trained individually. These results suggest that layer mapping in the same models, which are affected by the DSN problem, fails to alleviate the DSN problem within the teacher model, and highlight that the compact student is essential for effectively resolving DSN problem in CLDA, as shown in `M3'.
\vspace{-0.5cm}

\paragraph{Layer-wise relation. } {In Fig.~\ref{experiments:fig5} (b), we verified why updating the non-salient layer of the teacher to the most similar layer of the student is reasonable. We first computed the similarity between non-salient layers at the same position in the teacher and student models of the same size in `M2'. In `M3', we also computed the similarity between the teacher's non-salient layer and the most similar student layer. The results indicate that the layer similarity of `M3' exhibits lower similarity and greater variability compared to the similarity between non-salient layers at the same location of `M2'.
This indicates that the student layers demonstrate representational diversity and play a complementary role in refining non-salient layers of the teacher model.}

\begin{table}
\centering
\caption{Comparison of mIoU scores (\%) with existing KD-based UDA method on the GTA-to-Cityscapes and Synthia-to-Cityscapes tasks.}
\vspace{-0.2cm}
\small\addtolength{\tabcolsep}{4pt}
\begin{adjustbox}{max width=\linewidth} 
\begin{tabular}{l|c|c|c|c}
\toprule
\multirow{2}{*}{Method} & \multicolumn{2}{c|}{GTA$\rightarrow$City} & \multicolumn{2}{c}{Synthia$\rightarrow$City} \\
\cmidrule(lr){2-3} \cmidrule(lr){4-5}
& Teacher & Student & Teacher & Student \\
\midrule
\cite{kothandaraman2021domain} & 68.3  &  69.0  & 60.9 & 60.3 \\
\rowcolor{gray!20} Ours & 69.0 & 69.5 & 61.7 & 61.4 \\
\bottomrule
\end{tabular}
\end{adjustbox} 
\label{table:KD}
\vspace{-0.3cm}
\end{table}


\vspace{-0.4cm}
\paragraph{Comparison with KD method. }
As CLDA is built upon a distillation-based approach, we further compared with existing distillation-based UDA approaches such as~\cite{kothandaraman2021domain} in Table~\ref{table:KD}. However, we found it unfair to directly compare our method based on Transformer with~\cite{kothandaraman2021domain}, which is based on CNN. To ensure a fair comparison, we re-implemented \cite{kothandaraman2021domain} using DAFomer and conducted experiments on GTA-to-Cityscapes and Synthia-to-Cityscapes. Our method outperforms the previous approach in both datasets. A key distinction is that while the previous approach~\cite{kothandaraman2021domain} employs a fixed teacher model to train the student, our method leverages the collaborative learning strategy, where the teacher model is progressively updated during training. This not only improves the performance of the teacher model, but also improves the performance of the student model by transferring enhanced teacher knowledge to the student model. 

\vspace{-0.4cm}
\paragraph{Computational overhead.}
Here, we discuss the additional computational overhead introduced in the process of establishing layer-wise relations. Overall, we believe that the layer-wise computation cost is at an acceptable level. The layer-wise relations $T_{LR}$ are built in only 1\% of the total training iterations, resulting in just a 0.13\% increase in training complexity, which has minimal impact on computational cost. Furthermore, this overhead occurs only during training and does not introduce any additional cost during inference.

\section{Conclusion}
\label{sec:conclusion}

We explore the underexamined yet practical challenge of developing compact and efficient models in DA. We analyze the DSN issue caused by a fixed teacher model in conventional KD within DA. Based on this analysis, we propose CLDA, a practical approach that addresses the DSN problem by leveraging the complementary effects between teacher and student models. A notable strength of CLDA is its ability to resolve the DSN issue without additional backpropagation for the teacher model. Our empirical results demonstrate significant performance gains for both teacher and student models across various settings. Future work could include extending the proposed methodology to broader applications, such as domain generalization.

{\small \bibliographystyle{ieeenat_fullname}
    \bibliography{main}}

\end{document}